%% file: main.tex
\documentclass[10pt,twocolumn,letterpaper]{article}

\usepackage{iccv}    
\usepackage{algorithm}
\usepackage{algpseudocode} 
\usepackage{multirow}
\usepackage{bbm}

\newcommand{\vect}[1]{\mathbf{#1}} % For bold lowercase vectors
\newcommand{\mat}[1]{\mathbf{#1}}  % For bold uppercase matrices

\definecolor{iccvblue}{rgb}{0.21,0.49,0.74}
\usepackage[pagebackref,breaklinks,colorlinks,allcolors=iccvblue]{hyperref}

\def\confName{ICCV}
\def\confYear{2025}

\title{PLOT-TAL: Prompt-Learning with Optimal Transport for Few-Shot Temporal Action Localization}

\author{Edward Fish\\
University of Surrey \\
{\tt\small edward.fish@surrey.ac.uk}
\and
Andrew Gilbert\\
University of Surrey\\
{\tt\small a.gilbert@surrey.ac.uk}
}

\begin{document}
\maketitle
\input{sec/0_abstract}    
\input{sec/1_intro}
\input{sec/2_literature_review}

\input{sec/3_methodology}
\input{sec/4_results}

\input{sec/5_conclusion}
{
    \small
    \bibliographystyle{ieeenat_fullname}
    \bibliography{main}
}

\end{document}

%% file: sec/0_abstract.tex
\begin{abstract}
\noindent Few-shot temporal action localization (TAL) methods that adapt large models via single-prompt tuning often fail to produce precise temporal boundaries. This stems from the model learning a non-discriminative mean representation of an action from sparse data, which compromises generalization. We address this by proposing a new paradigm based on multi-prompt ensembles, where a set of diverse, learnable prompts for each action is encouraged to specialize on compositional sub-events. To enforce this specialization, we introduce PLOT-TAL, a framework that leverages Optimal Transport (OT) to find a globally optimal alignment between the prompt ensemble and the video's temporal features. Our method establishes a new state-of-the-art on the challenging few-shot benchmarks of THUMOS'14 and EPIC-Kitchens, without requiring complex meta-learning. The significant performance gains, particularly at high IoU thresholds, validate our hypothesis and demonstrate the superiority of learning distributed, compositional representations for precise temporal localization.

\end{abstract}

%% file: sec/1_intro.tex
\section{Introduction}
\label{sec:intro}

Temporal Action Localization (TAL) is the task of identifying the start, end, and class labels of actions in continuous videos. While the success of state-of-the-art models has been predicated on access to vast, densely annotated datasets, for TAL to be deployed robustly in real-world applications where data is inherently scarce, these networks need to be able to efficiently learn from only a few samples.

\begin{figure*}[t]
    \centering
    % In a real paper, you would replace this with the actual figure file.
\includegraphics[width=\linewidth]{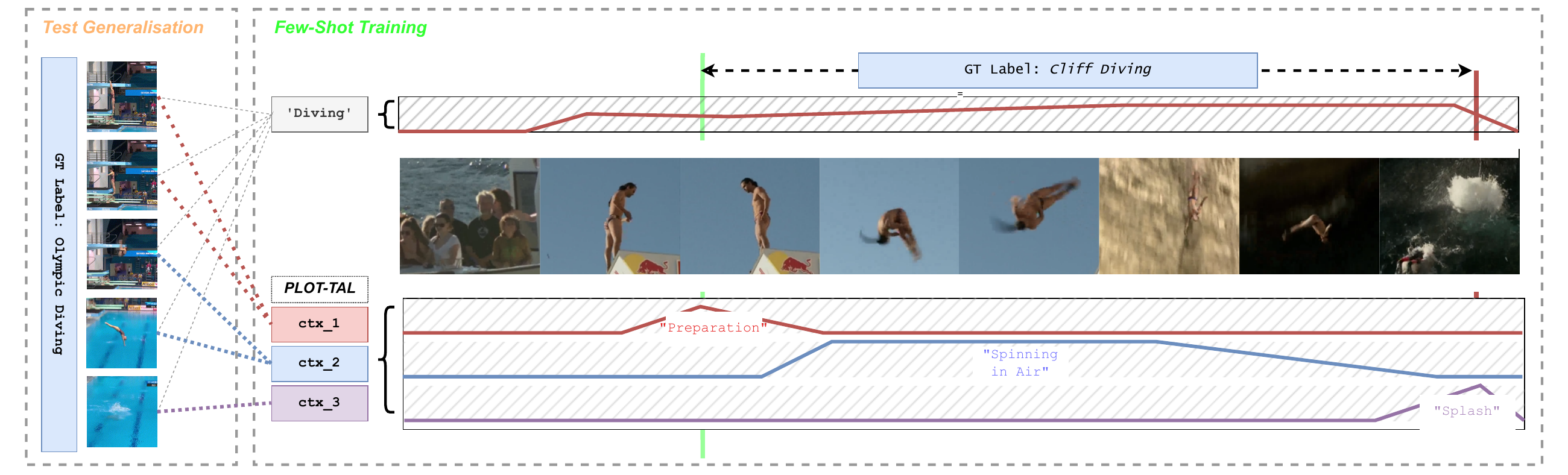} 
\caption{\textbf{Conceptual Framework: Compositional Learning for Few-Shot Generalization.} 
A single prompt trained on a few examples of ``diving'' in a specific context (top-right) tends to overfit to environmental cues like the cliffs and sea. This holistic representation fails to generalize to a novel environment. In contrast, our method (bottom right) learns an ensemble of prompts that specialize on the compositional, environment-agnostic sub-events of the action: (1) the preparation/stance, (2) the mid-air rotation, and (3) the water entry splash. Optimal Transport is the key mechanism that enforces this specialization, ensuring the prompts remain diverse and discriminative. By identifying these core components, our framework can robustly localize the ``diving'' action with high precision, even when presented with a completely different environment, such as an indoor swimming pool (left panel), using only a few samples.}
\label{fig:overview}
\end{figure*}

The current strategy for tackling this low-data regime involves adapting large Vision-Language Models (VLMs) \cite{radford2021learning} via parameter-efficient prompt tuning \cite{zhou2022learning}. However, the limitations of this paradigm become critically apparent in the few-shot setting. By learning a single prompt per class, the model is forced to compress the entire dynamic structure of an action into a single feature vector. This representational bottleneck is severely exacerbated when learning from limited samples. With only a handful of examples, a single prompt is prone to memorizing superficial, non-generalizable cues (e.g., the specific camera angle or background clutter) present in the few shots, leading to poor generalization to novel contexts.

In this work, we propose that robust few-shot generalization stems not from learning a single, monolithic concept, but from discovering the underlying compositional structure of actions. For example, an action like a `high jump' is a composition of simpler, more reusable sub-events (`running', `leaping', `arching the back'). Learning these disentangled concepts from sparse data is a more tractable problem. Consequently, we depart from the single-prompt paradigm and propose modelling each action class with a set of diverse, learnable prompts.

This approach requires a mechanism to guide the specialization of these prompts and prevent their collapse into redundancy. For this, we introduce Optimal Transport (OT) \cite{cuturi2013sinkhorn}, not merely as a matching algorithm, but as a structural regularizer that enforces representational diversity. By seeking the most efficient assignment between the distribution of prompts and the distribution of temporal features, OT implicitly ensures that each prompt finds a unique role in explaining the data. This constraint is a powerful tool against overfitting, preventing the entire set of prompts from aligning with the most prominent feature in the few training examples, thereby fostering a diverse and highly generalizable final representation.

\noindent
\textbf{Our contributions are as follows:}

\begin{itemize}
    \item We identify the single-prompt architecture as a key source of poor generalization in few-shot TAL.
    \item We propose a multi-prompt, OT-regulated framework as a direct solution, arguing it learns a more compositional representation inherently better suited to low-data regimes.
    \item We provide extensive empirical validation, demonstrating state-of-the-art performance on multiple benchmarks.
\end{itemize} 

%% file: sec/2_literature_review.tex
\section{Related Work}
\label{lit-review}

The field of TAL has evolved from two-stage methods, which first generate proposals and then classify them \cite{escorcia2016daps, lin2018bsn, lin2019bmn}, towards more efficient single-stage architectures. These unified models, inspired by advances in object detection \cite{redmon2016you, liu2016ssd}, perform classification and boundary regression in a single pass. Recent state-of-the-art methods frequently leverage Transformer-based backbones \cite{vaswani2017attention} combined with feature pyramid networks (FPNs) to handle actions at multiple temporal scales \cite{zhang2022actionformer, cheng2022tallformer, shi2023temporal}. While powerful, these models' performance relies on large-scale supervision, a limitation our work aims to address.

\subsection{Few-Shot Learning for Action Localization}
Adapting TAL to the few-shot setting has primarily been explored through meta-learning \cite{yang2018one, xu2020revisiting}. These methods train a model to quickly adapt to novel classes by learning across a distribution of tasks, or ``episodes.'' While effective, they often involve complex, multi-stage training schedules. Other approaches have tackled zero-shot TAL \cite{nag2022zero}, but often rely on external cues from pre-trained classifiers like UntrimmedNet \cite{wang2017untrimmed}, which may not be available in practical scenarios. Our approach provides a simpler, end-to-end framework for few-shot learning that circumvents complex meta-learning and external dependencies.

\subsection{Prompt Learning in Vision}
Prompt learning has emerged as a parameter-efficient method for adapting large, frozen VLMs to new tasks \cite{zhou2022learning, zhou2022conditional}. By only tuning a small number of context vectors prepended to a text prompt, these methods can steer the model's behaviour without updating all of its parameters. This has been applied to action recognition \cite{ju2022prompting} and video-text alignment \cite{li2022align}, but its application to the fine-grained task of localization remains under explored. Works like \cite{nag2022multi} have combined prompting with meta-learning, but our work focuses on a more direct adaptation for few-shot TAL and extends to multiple prompts.

\subsection{Optimal Transport in Machine Learning}
The Optimal Transport (OT) problem, originating from the work of Monge \cite{villani2009optimal}, provides a principled way to measure the distance between probability distributions. Its applicability to machine learning was greatly expanded by the introduction of entropic regularization and the efficient Sinkhorn algorithm \cite{cuturi2013sinkhorn}, which made it computationally tractable for high-dimensional data. OT has since been applied to various vision tasks \cite{torres2021survey}. Most relevant to our work is Chen et al. \cite{chen2022prompt}, who first used OT to align multiple prompts to feature maps for few-shot image classification. Our primary contribution is the novel adaptation and validation of this concept for the temporal domain, demonstrating that OT is a powerful tool for modelling the dynamic, compositional structure of actions over time—a fundamentally different and more complex problem than static image classification.

%% file: sec/3_methodology.tex
\section{Methodology}
\label{sec:methodology}

\begin{figure}[t]
    \centering
    % In a real paper, you would replace this with the actual figure file.
    \includegraphics[width=\columnwidth]{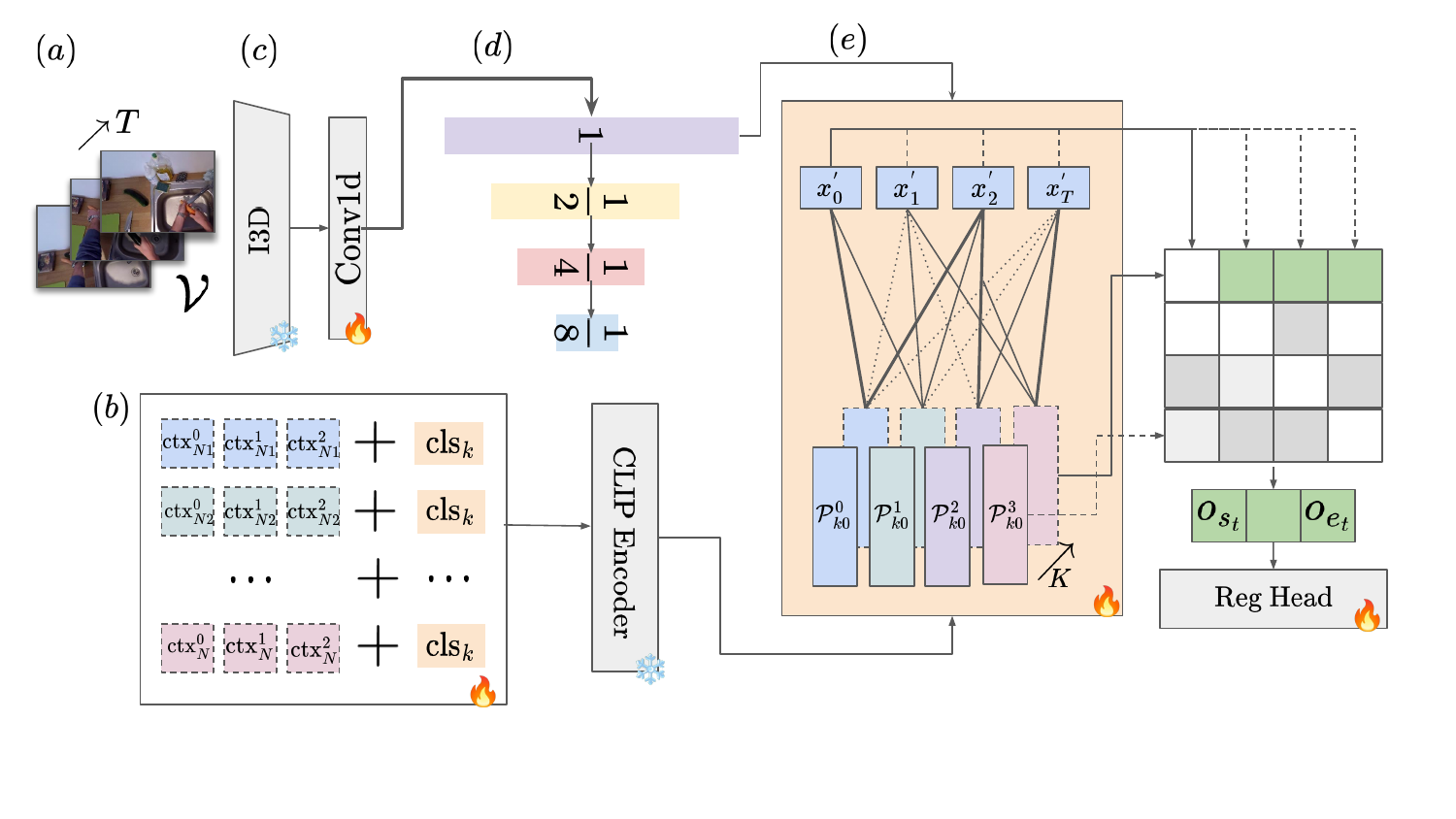} 
    \caption{An overview of the PLOT-TAL framework. \textbf{(A)} We first extract $T$ frames from a video $V$. \textbf{(B)} An ensemble of $N$ learnable prompts is generated for each class. \textbf{(C)} Video features are extracted by a frozen visual encoder and text prompts by a frozen VLM text encoder (CLIP). \textbf{(D)} A temporal feature pyramid is constructed via max-pooling. \textbf{(E)} Optimal Transport aligns the prompt ensemble with video features at each pyramid level. \textbf{(F)} The resulting features are passed to lightweight localization heads to predict action instances and a class label. Only modules marked with a flame symbol contain trainable parameters.}
    \label{fig:method}
\end{figure}

We propose a novel framework for Temporal Action Localization (TAL), which we term PLOT-TAL. Our approach integrates pre-trained feature extraction, adaptive multi-prompt learning, and an efficient feature-prompt alignment mechanism based on the Sinkhorn algorithm for Optimal Transport. The framework is designed to be parameter-efficient and is trained end-to-end. An overview of the network architecture is presented in Figure \ref{fig:method}.

\subsection{Problem Formulation}
We first formally define the task. Given an untrimmed input video, represented as a sequence of feature vectors $\mathcal{V} = \{\vect{v}_1, \vect{v}_2, \dots, \vect{v}_T\}$, the objective of TAL is to predict a set of action instances $\mathcal{Y} = \{(s_i, e_i, c_i)\}_{i=1}^M$. Each tuple denotes an action of class $c_i \in \{1, \dots, C\}$ starting at time $s_i$ and ending at time $e_i$, where the temporal boundaries must satisfy $1 \le s_i < e_i \le T$.

In the few-shot setting this paper addresses, the model must learn to perform this task for all $C$ classes using only a small number, $K$, of annotated support examples for each class (e.g., $K=5$). This constraint requires a model that can generalize effectively from sparse data while minimizing the number of trainable parameters to prevent overfitting.

\subsection{Multi-Scale Feature and Prompt Representation}
Our framework is designed to capture actions at varying temporal scales by using hierarchical representations for both visual features and textual prompts.

\subsubsection{Temporal Feature Pyramid}
The input video is first processed by a frozen, pre-trained 3D Convolutional Network (e.g., I3D \cite{carreira2017quo}) to extract a sequence of clip-level feature vectors $\{\vect{x}_1, \dots, \vect{x}_T\}$. To enhance local temporal context, this sequence is passed through a series of 1D temporal convolutional layers, following modern TAL architectures \cite{zhang2022actionformer,shi2023tridet}. This yields a refined feature sequence $\{\vect{x'}_1, \dots, \vect{x'}_T\}$. From this base sequence, we construct a temporal feature pyramid of $L$ levels by applying successive max-pooling operations with a stride of 2. This results in a set of multi-scale feature representations $\{\mat{F}_1, \dots, \mat{F}_L\}$, where each matrix $\mat{F}_l \in \mathbb{R}^{T_l \times D}$ contains the feature sequence at temporal scale $l$.

\subsubsection{Adaptive Multi-Prompt Ensembles}
To address the limitations of a single-prompt representation, we model each action class $c$ with an ensemble of $N$ diverse, learnable prompts. Each of the $N$ prompts is constructed by prepending a unique set of $n_{\text{ctx}}$ learnable context vectors to the class name: $[\text{ctx}_1, \dots, \text{ctx}_{n_{\text{ctx}}}, \text{class\_name}_c]$. This set of $N$ textual prompts is then passed through the frozen text encoder of a pre-trained VLM like CLIP \cite{radford2021learning}. The process is formalized as:
\begin{equation}
    \vect{g}_{ci} = f_{\text{CLIP}}(\text{class\_name}_c, \{\text{ctx}_{cij}\}_{j=1}^{n_{\text{ctx}}})
\end{equation}
where $\vect{g}_{ci}$ is the $i$-th prompt embedding for class $c$. This yields a set of prompt embeddings $\mat{G}_c = \{\vect{g}_{c1}, \dots, \vect{g}_{cN}\} \in \mathbb{R}^{N \times D}$. The only trainable parameters in this module are the context vectors $\{\text{ctx}\}$, ensuring our approach remains highly parameter-efficient.

\subsection{Multi-Resolution Alignment via Optimal Transport}
The core technical contribution of our work is the mechanism for aligning the prompt ensemble $\mat{G}_c$ with the temporal feature sequence $\mat{F}_l$ at each pyramid level $l$.

\subsubsection{Optimal Transport Formulation} We formulate the alignment as a distribution matching problem. For a given class $c$ and level $l$, we treat the set of $T_l$ video features and $N$ prompt embeddings as empirical samples from two discrete probability distributions, $U_l$ and $V_c$, respectively:
\begin{equation}
    U_l = \sum_{i=1}^{T_l} u_i \delta_{\vect{f}_i} \quad \text{and} \quad V_c = \sum_{j=1}^{N} v_j \delta_{\vect{g}_j}
\end{equation}
where $\delta_{\cdot}$ is the Dirac delta function, and $\vect{u}$ and $\vect{v}$ are uniform probability vectors (i.e., $u_i = 1/T_l$, $v_j = 1/N$). We define a cost matrix $\mat{C} \in \mathbb{R}^{T_l \times N}$ where each entry $C_{ij}$ is the cosine distance between the video feature $\vect{f}_i$ and the prompt embedding $\vect{g}_j$. The goal of OT is to find a transport plan $\mat{T} \in \mathbb{R}^{T_l \times N}$ that minimizes the total transportation cost. We use the entropically regularized formulation, solvable efficiently via the Sinkhorn algorithm \cite{cuturi2013sinkhorn}:
\begin{equation}
\label{eq:ot}
d_{\text{OT}}(\mat{F}_l, \mat{G}_c) = \min_{\mat{T} \in \mathcal{U}(\vect{u}, \vect{v})} \langle \mat{T}, \mat{C} \rangle - \lambda H(\mat{T})
\end{equation}
Here, $\langle \cdot, \cdot \rangle$ is the Frobenius dot product, $H(\mat{T})$ is the entropy of the transport plan, and $\lambda$ is a regularization parameter. The resulting optimal transport plan $\mat{T}^*$ represents a soft, many-to-many assignment map.

\subsubsection{Optimization} This alignment process is embedded in a two-stage optimization loop as proposed in \cite{chen2022prompt}. In the inner loop of each training step, we fix the model parameters and iteratively solve Eq. \ref{eq:ot} to find the optimal transport plan $\mat{T}^*_l$. In the outer loop, with the transport plans fixed, we compute the final task loss and backpropagate the gradients through the OT process to update the learnable prompt context vectors. For the OT-specific parameters, we follow the setup in \cite{chen2022prompt}, setting the convergence threshold $\delta = 0.01$, the entropy parameter $\lambda = 0.1$, and we perform a maximum of 100 iterations within the inner Sinkhorn loop. A detailed overview of this process is provided in Algorithm \ref{alg:detailedOTsinkhorn}.

\begin{algorithm}[t]
\caption{PLOT-TAL Optimization Loop}
\label{alg:detailedOTsinkhorn}
\begin{algorithmic}[1]
\State \textbf{Input:} Video features $\{\mat{F}_l\}_{l=1}^L$, class labels $\{c\}$
\State \textbf{Output:} Optimized context vectors $\{\text{ctx}\}$
\State Initialize learnable context vectors $\{\text{ctx}\}$
\For{each training iteration}
    \For{each class $c$ and pyramid level $l$}
        \State Generate prompt embeddings $\mat{G}_c \in \mathbb{R}^{N \times D}$
        \State Calculate cost matrix $\mat{C}_{l,c} = 1 - \mat{F}_l \mat{G}_c^\top$
        \State \textit{//--- Inner Loop: Sinkhorn Algorithm ---}
        \State Initialize $\vect{v} \leftarrow \mathbf{1}/N$
        \For{$t_{in} = 1$ to $T_{in}$}
            \State $\vect{u} \leftarrow \mathbf{1} / (\exp(-\mat{C}_{l,c}/\lambda) \vect{v})$
            \State $\vect{v} \leftarrow \mathbf{1} / (\exp(-\mat{C}_{l,c}/\lambda)^\top \vect{u})$
        \EndFor
        \State Compute transport plan $\mat{T}^*_{l,c}$ from $\vect{u}, \vect{v}$
        \State Compute OT distance $d_{\text{OT}}(l,c) = \langle \mat{T}^*_{l,c}, \mat{C}_{l,c} \rangle$
    \EndFor
    \State \textit{//--- Outer Loop ---}
    \State Compute final predictions using aligned features
    \State Compute total loss $\mathcal{L}_{\text{total}}$ (Eq. 4)
    \State Backpropagate gradients from $\mathcal{L}_{\text{total}}$ to update $\{\text{ctx}\}$
\EndFor
\State \textbf{return} Optimized context vectors $\{\text{ctx}\}$
\end{algorithmic}
\end{algorithm}

\subsection{Decoder and Learning Objective}
\subsubsection{Decoder Architecture.}
Following the multi-scale alignment, the resulting features are passed to two lightweight, parallel heads for the final predictions: a classification head that generates a probability distribution over the $C$ classes using a sigmoid activation, and a regression head that predicts the temporal offsets to the start and end boundaries of a potential action using a ReLU activation.

\subsubsection{Learning Objective}
The network is trained end-to-end by minimizing a total loss function, $\mathcal{L}_{\text{total}}$. We use the Focal Loss ($\mathcal{L}_{\text{cls}}$) \cite{lin2017focal} for classification and the Distance-IoU (DIoU) Loss ($\mathcal{L}_{\text{reg}}$) \cite{zheng2020distance} for regression. The total loss, aggregated over all temporal locations $t$ and pyramid levels $l$, is:
\begin{equation}
\label{eq:total_loss_split} % A new label for the revised equation
\begin{split}
    \mathcal{L}_{\text{total}} = \frac{1}{N_{pos}} \sum_{l,t} \Big( & \mathcal{L}_{\text{cls}}(\hat{c}_{lt}, c_{lt}) \\
    & + \lambda_{reg} \mathbbm{1}_{\{c_{lt}>0\}} \mathcal{L}_{\text{reg}}(\hat{o}_{lt}, o_{lt}) \Big)
\end{split}
\end{equation}
where $\hat{c}_{lt}$ and $\hat{o}_{lt}$ are the predictions, $N_{pos}$ is the number of positive samples, and the indicator function $\mathbbm{1}_{\{\cdot\}}$ applies the regression loss only to foreground frames.

%% file: sec/4_results.tex
\section{Experiments}
\label{sec:experiments}

We conduct a comprehensive set of experiments to rigorously validate our proposed framework, PLOT-TAL.
\subsection{Experimental Setup}

\subsubsection{Datasets and Metrics}
Our evaluation is performed on two standard, yet diverse, benchmarks for temporal action localization:
\begin{itemize}
    \item \textbf{THUMOS’14} \cite{THUMOS14} is a widely used dataset featuring 20 classes of sports actions in 200 validation and 213 test videos.
    \item \textbf{EPIC-Kitchens-100} \cite{damen2018scaling} is a large-scale egocentric dataset. We evaluate on both the verb (97 classes) and noun (300 classes) localization tasks.
\end{itemize}
Following standard protocols, we report the mean Average Precision (mAP) at various Intersection over Union (IoU) thresholds: [0.3, 0.4, 0.5, 0.6, 0.7], and report the average of these as our primary metric.

\subsubsection{Few-Shot Protocol and Comparison to Prior Work}
All our experiments are conducted under a 5-shot, C-way protocol, where $C$ is the total number of classes in the respective dataset. In this challenging setup, the model learns from only 5 examples per class and must then localize actions from among all $C$ classes simultaneously during testing. This differs significantly from the episodic 5-shot, 5-way meta-learning protocol used in some prior works \cite{nag2021few, keisham2023multi}. Due to this fundamental difference in task difficulty, results from those works are presented for context but are not directly comparable.

\subsubsection{Baselines}
We compare PLOT-TAL against three strong baselines trained under the exact same few-shot protocol for a fair comparison:
\begin{enumerate}
    \item \textbf{Linear Probe (LP)}: A linear classifier trained on top of the frozen video features.
    \item \textbf{CoOp} \cite{zhou2022learning}: The canonical single-prompt learning method.
    \item \textbf{Ours (Avg.)}: An ablation of our model where the $N$ prompts are combined by simple averaging, removing the Optimal Transport module.
\end{enumerate}

\subsubsection{Implementation Details}
We use standard I3D (RGB+Flow) features for THUMOS’14 and SlowFast for EPIC-Kitchens. Models are trained for 100 epochs using the Adam optimizer with a batch size of 2 on a single NVIDIA RTX 3090 GPU. Based on our ablation studies, we set the number of prompts $N=6$ and context tokens $n_{ctx}=16$. The OT regularization $\lambda$ is set to $0.1$ following \cite{chen2022prompt}. All results are averaged over 4 random seeds.

\begin{table*}
\scriptsize
\centering
\caption{Performance comparison of our proposed method PLOT-TAL on the THUMOS-14 dataset against baselines.}
\label{tab:thumos_comparison}
\begin{tabular}{@{}lccccc|c@{}}
\toprule
Method & mAP@0.3 & mAP@0.4 & mAP@0.5 & mAP@0.6 & mAP@0.7 & Avg (mAP) \\ \midrule
Baseline I (avg) & 37.3 & 32.93 & 26.88 & 18.17 & 8.83 & 24.82 \\
Baseline II (lp) & 51.98 & 46.5 & 36.79 & 25.62 & 14.66 & 35.11 \\
CoOP & 48.73 & 43.67 & 36.64 & 27.24 & 16.97 & 34.65 \\
\midrule
PLOT-TAL CLS & 53.46 & 48.93 & 38.2 & 30.2 & 18.8 & 38.24 \\
PLOT-TAL Verbose & \textbf{56.42} & \textbf{50.54} & \textbf{42.48} & \textbf{32.35} & \textbf{21.17} & \textbf{40.59} \\ \bottomrule
\end{tabular}

\end{table*}

\subsection{Evaluation}
This section evaluates our approach against existing methods for both few-shot temporal action localisation and prompt learning. To compare with previous works, we report the mean average precision (mAP) at various intersections over union for all results.

\subsubsection{THUMOS-14} 
In Tab~\ref{tab:thumos_comparison}, we show results for 5-shot 20-way TAL on the THUMOS'14 dataset for our approach \textit{PLOT-TAL CLS}. Adding additional class prompts can improve performance over a single prompt by a large margin ($\uparrow5.9$). We also show how it's possible to achieve higher accuracy by handcrafting prompts (Verbose). In this setting, we use GPT-3.5 \cite{brown2020language} to produce additional descriptions of the actions that will replace the class label.

The Baseline $I$ method represents performance when we add additional prompts but exclude optimal transport, demonstrating how optimal transport is highly effective at aligning the features ($\uparrow15.77$). While Baseline $II$ (linear probe) based on the work of \cite{radford2021learning} and \cite{chen2022prompt} has an average performance of 5\% less than our method.

\begin{figure}
\centering
\includegraphics[width=\linewidth]{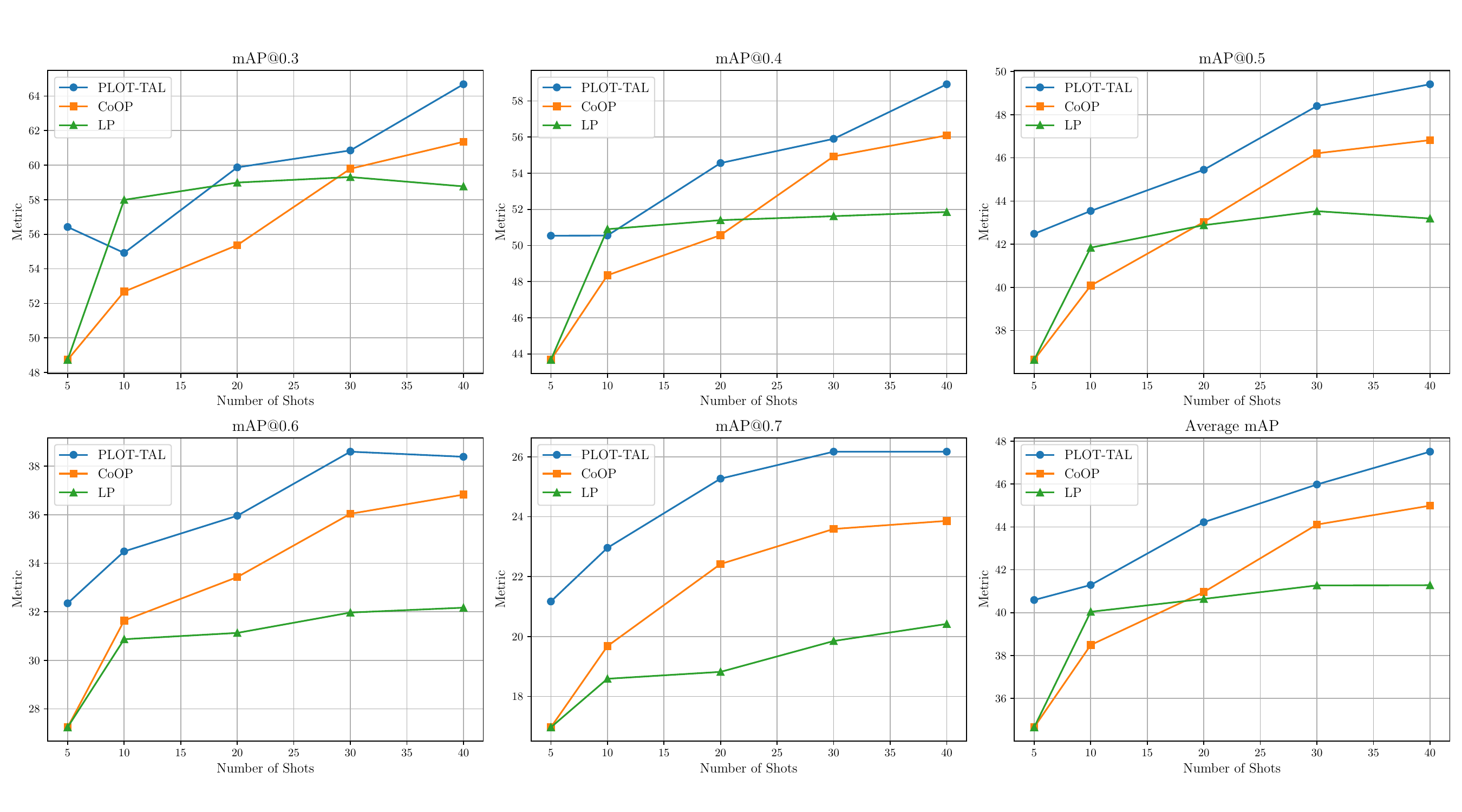}
\caption{mAP over various IOU thresholds and number of training samples on the THUMOS-14 dataset. Additional prompts demonstrate improved performance, especially at high IoU thresholds indicating improved discriminative ability.}
\label{fig:mapIoU}
\end{figure}

In Fig~\ref{fig:mapIoU}, we demonstrate how the optimal transport improves performance at higher IoU thresholds than single prompt or linear probe methods.

At low IoU thresholds, the predicted segment only needs to overlap with a small section of the ground truth, meaning that single prompt methods and linear probes achieve relatively good performance as they distribute the attention between prompts and features across the temporal domain. However, as we increase the IoU threshold, we can see that our PLOT-TAL method becomes more effective, demonstrating the network's higher discriminative ability. 

\subsubsection{EPIC-KITCHENS-100}

In Tab~\ref{tab:epic_kitchens_revised}, we show results on the EPIC Kitchens verb and noun partitions, showing a slight improvement over single prompt methods for the noun classes ($\uparrow1.19$) but achieve a more significant performance boost for the verb classes ($\uparrow2.96$). 

\begin{table*}[t]
\centering
\caption{Few-shot TAL performance (mAP@IoU) on the EPIC-Kitchens-100 Noun and Verb partitions. Our method shows the most significant gains on the more dynamic verb classes, supporting our thesis that it excels at modeling complex temporal structure.}
\label{tab:epic_kitchens_revised}
\small
\begin{tabular}{@{}l ccccc c ccccc c@{}}
\toprule
& \multicolumn{6}{c}{\textbf{EPIC-Kitchens Noun}} & \multicolumn{6}{c}{\textbf{EPIC-Kitchens Verb}} \\
\cmidrule(lr){2-7} \cmidrule(lr){8-13}
\textbf{Method} & \textbf{@0.1} & \textbf{@0.2} & \textbf{@0.3} & \textbf{@0.4} & \textbf{@0.5} & \textbf{Avg.} & \textbf{@0.1} & \textbf{@0.2} & \textbf{@0.3} & \textbf{@0.4} & \textbf{@0.5} & \textbf{Avg.} \\
\midrule
Ours (Avg.) & 14.3 & 13.5 & 13.1 & 10.3 & 9.3 & 12.1 & 21.2 & 19.9 & 18.0 & 15.2 & 11.9 & 17.3 \\
Linear Probe (LP) & \textbf{18.0} & 15.4 & 14.1 & 12.2 & 9.5 & 13.9 & \textbf{22.5} & \textbf{21.3} & 19.2 & 17.1 & 13.3 & 18.7 \\
CoOp \cite{zhou2022learning} & 16.1 & 15.0 & 13.8 & 11.8 & 9.5 & 13.3 & 18.5 & 17.6 & 16.3 & 14.6 & 12.5 & 15.9 \\
\midrule
\textbf{PLOT-TAL (Ours)} & 17.9 & \textbf{16.7} & \textbf{15.1} & \textbf{12.7} & \textbf{10.0} & \textbf{14.5} & 21.8 & 20.9 & \textbf{19.4} & \textbf{17.6} & \textbf{14.6} & \textbf{18.9} \\
\bottomrule
\end{tabular}
\end{table*}

\begin{table}[t]
\centering
\small
% The caption is now at the top, and the label is correctly placed after it.
\caption{Few-shot TAL performance on THUMOS'14. Our end-to-end (E2E) method is compared against prior meta-learning (ML) work. Note that the evaluation settings are different, making a direct comparison of scores challenging; our 20-way task is significantly harder than the 5-way task.}
\label{tab:results-few-shot-thumos-revised}
\begin{tabular}{@{}llc@{}}
\toprule
\textbf{Method} & \textbf{Approach} & \textbf{Avg. mAP (\%)} \\
\midrule
\multicolumn{3}{l}{\textit{Meta-Learning Approaches (5-shot, 5-way)}} \\ 
\cmidrule(r){1-3} 
Common Action Loc. \cite{yang2020localizing} & ML & 22.8 \\
MUPPET \cite{nag2022multi} & ML + PL & 24.9 \\
Multi-Level Align. \cite{keisham2023multi} & ML & 31.8 \\
Q. A. Transformer \cite{nag2021few} & ML & 32.7 \\
\midrule
\multicolumn{3}{l}{\textit{End-to-End Prompt Learning (5-shot, 20-way)}} \\
\cmidrule(r){1-3}
CoOp \cite{zhou2022learning} & E2E + PL & 34.65 \\
\textbf{PLOT-TAL (Ours)} & E2E + PL & \textbf{38.24} \\
\textbf{PLOT-TAL (Verbose) (Ours)} & E2E + PL & \textbf{40.59} \\
\bottomrule
\end{tabular}
\end{table}

This demonstrates the effectiveness of the additional prompts in distinguishing between complex temporal features. However, the performance improvement is less pronounced for the noun partition. This suggests that nouns, which are generally static and visually distinct, are inherently easier to classify with a single prompt. As a result, they do not derive as much benefit from the added context provided by multiple prompts. Nouns typically represent objects with consistent visual appearances, reducing the need for additional context to disambiguate them. Therefore, the application of optimal transport, which excels in aligning distributions of more dynamic and context-dependent features (such as verbs), does not yield a substantial advantage in this case.

In Tab~\ref{tab:results-few-shot-thumos-revised}, we compare with other SOTA methods for few-shot temporal action localisation, which utilise meta-learning and perform few-shot localisation at a $5$-shot, $5$-way setting, whereas our results are from the $5$-shot, $20$-way configuration. Not only is the $5$-shot, $20$-way few-shot setting more challenging, but PLOT-TAL also benefits from being trained end-to-end without the requirement for pre-training and episodic adaptive contrastive learning as in current meta-learning approaches.

\subsection{Qualitative Results}

\begin{figure}[t]
\centering
\includegraphics[width=\linewidth]{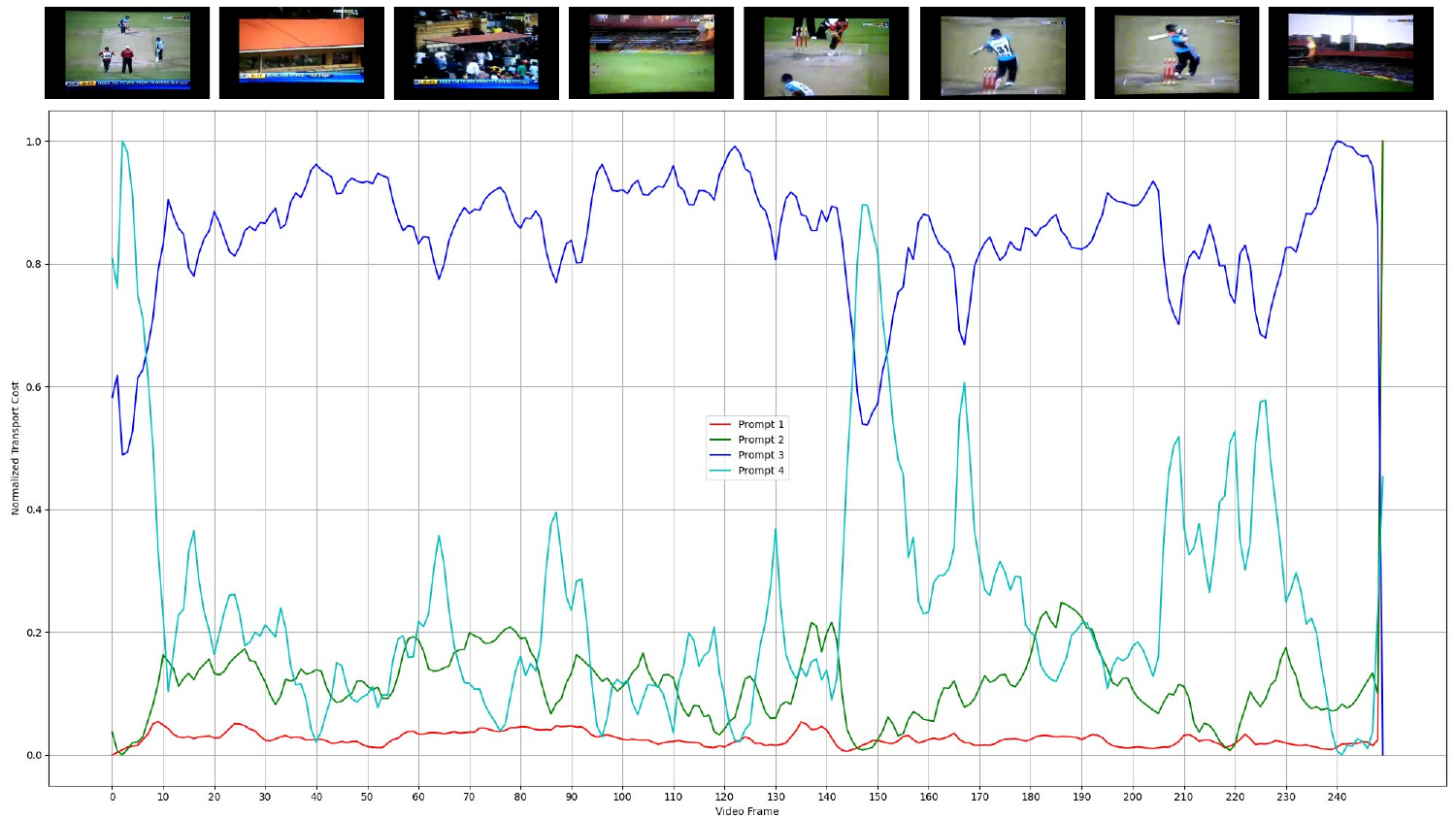}
\caption[Transport cost for each $N$ prompt for the class `Cricket Shot'.]{The normalised transport cost of each $N$ prompt for the class `Cricket Shot' after training. Prompt one aligns with global information, while the other prompts learn additional, complementary views. In the transport cost algorithm, a lower value indicates closer alignment. }
\label{fig:t_plot}
\end{figure}

In Fig~\ref{fig:t_plot}, we show the normalised transport cost for each frame and $N$ embedding for the class label `Cricket Shot'. This figure shows how each $N$ prompts diverge and focuses on different elements and views within the videos. For example, we can see that $N_1$ or Prompt 1 learns global information across all frames. This shows how we may distribute alignment across all frames in a single prompt framework and lose discriminative ability since it learns global information over the whole video. In the figure, we can note that Prompt 4 appears to learn background information and is more closely aligned to frames where we can see the stadium stands. Prompts 2 and 3, however, indicate a closer alignment with objects related to the class of `cricket shot,' including when the cricket strip is in the shot and there are people on the field. The plot clearly reveals that each prompt, when considered in isolation, demonstrates varying degrees of alignment with different sections of the video. This suggests that each prompt is capturing unique and complementary aspects of the video content, allowing for a more nuanced understanding of the temporal dynamics.

\subsection{Ablation Experiments}
We perform several ablation experiments to evaluate each component of the architecture. We experiment with the number of learnable context tokens and prompts per class, alternative feature alignment metrics, and the number of feature pyramid network levels. We also experimented with the types of RGB embeddings and several prompt-engineering strategies.

\subsubsection{Number of Learnable Prompts}
In Tab~\ref{tab:n_ablation_revised}, we perform an ablation experiment on the number of learnable prompts $N$. The results show that the optimum number of prompts is $N=6$, while with an increased number of prompts, e.g., $N=10$, we can achieve better results in the more difficult IoU thresholds. This is due to the increased temporal discriminative ability of the additional prompts. As the N increases, performance degrades as the model overfits due to the increased number of learnable parameters. 

\begin{table}[t]
\centering
\small % Use a slightly smaller font size for dense tables
\caption{Ablation on the number of prompts ($N$) per class, evaluated on THUMOS'14. We report mAP (\%) at various IoU thresholds. Performance is optimal at $N=6$. Best result in each column is in bold.}
\label{tab:n_ablation_revised}
\begin{tabular}{@{}c ccccc c@{}}
\toprule
\multirow{2}{*}{\textbf{Prompts ($N$)}} & \multicolumn{5}{c}{\textbf{mAP @ IoU}} & \multirow{2}{*}{\textbf{Avg}} \\
\cmidrule(lr){2-6}
& \textbf{0.3} & \textbf{0.4} & \textbf{0.5} & \textbf{0.6} & \textbf{0.7} & \\
\midrule
4  & 55.88 & 50.21 & 43.06 & 31.97 & 21.16 & 40.46 \\
6  & \textbf{56.42} & \textbf{50.54} & 42.48 & 32.35 & 21.17 & \textbf{40.59} \\
8  & 53.60 & 48.72 & 41.74 & 31.68 & 20.70 & 39.29 \\
10 & 54.96 & 50.27 & \textbf{43.45} & \textbf{32.53} & \textbf{21.44} & 40.53 \\
12 & 53.74 & 48.25 & 41.02 & 30.57 & 20.06 & 38.73 \\
14 & 54.25 & 48.94 & 40.90 & 30.78 & 18.86 & 38.75 \\
16 & 53.66 & 48.28 & 41.04 & 30.84 & 20.15 & 38.79 \\
\bottomrule
\end{tabular}
\end{table}

\begin{table*}[t]
\centering
\small % Use a slightly smaller font to ensure it fits in a single column
\caption{Ablation on visual feature embeddings, evaluated on THUMOS'14. Combining RGB and Optical Flow (Flow) features from I3D yields the best performance, highlighting the importance of explicit motion cues for the TAL task.}
\label{tab:visual-embedding}
\begin{tabular}{@{}l ccccc c@{}}
\toprule
\multirow{2}{*}{\textbf{Embedding Type}} & \multicolumn{5}{c}{\textbf{mAP @ IoU}} & \multirow{2}{*}{\textbf{Avg. mAP (\%)}} \\
\cmidrule(lr){2-6}
& \textbf{0.3} & \textbf{0.4} & \textbf{0.5} & \textbf{0.6} & \textbf{0.7} & \\
\midrule
CLIP Vision (ViT-B-16) & 46.99 & 42.09 & 34.26 & 25.34 & 15.82 & 32.90 \\
RGB (I3D) & 43.13 & 38.76 & 31.71 & 23.15 & 14.46 & 30.24 \\
Optical Flow (I3D) & 26.03 & 23.10 & 19.54 & 14.07 & 8.93 & 18.33 \\
\midrule
\textbf{RGB + Flow (I3D)} & \textbf{55.88} & \textbf{50.21} & \textbf{43.06} & \textbf{31.97} & \textbf{21.16} & \textbf{40.46} \\
\bottomrule
\end{tabular}
\end{table*}
\subsubsection{Number of Learnable Context Tokens}

Each prompt also has several learnable context tokens as described in \cite{zhou2016learning} and \cite{weng2022efficient}. These context tokens are randomly initialised so that for the class `Basketball Dunk' with 4 $ctx$ tokens, the full prompt will be 
\begin{equation}
  {P} = {\{X, X, X, X, \text{Basketball Dunk}}\}
\end{equation}

%\begin{figure}[htbp]
%\centering
%\includegraphics[width=0.8\linewidth]{Paper4/Images/iou.png}
%\caption[Ablation experiment on the number of additional context tokens.]{mAP over various IoU thresholds for the THUMOS' 14 dataset with a variable number of additional context tokens appended to each $N$ prompt.}
%\label{fig:iou}
%\end{figure}

In Tab~\ref{tab:ctx}, we show the effect of varying the number of learnable $ctx$ tokens appended to each prompt. For each $N$ prompt, $n_{ctx}$ tokens are randomly initialised. The figure shows that the optimum number of tokens is between 10 and 20. As per the existing literature \cite{zhou2016learning, zhou2022conditional}, we select 16 tokens for all methods unless otherwise stated and train and test using the $5$-shot, $20$-way setup.
\begin{table}[t]
\centering
\small % Use a slightly smaller font size for dense tables
\caption{Ablation on the number of context tokens ($n_{\text{ctx}}$) per prompt, evaluated on THUMOS'14. We report mAP (\%) at various IoU thresholds. Performance is robust for values between 10-20, with the optimum at $n_{\text{ctx}}=16$.}
\label{tab:ctx}
\begin{tabular}{@{}c ccccc c@{}}
\toprule
\multirow{2}{*}{\textbf{$n_{\text{ctx}}$}} & \multicolumn{5}{c}{\textbf{mAP @ IoU}} & \multirow{2}{*}{\textbf{Avg.}} \\
\cmidrule(lr){2-6}
& \textbf{0.3} & \textbf{0.4} & \textbf{0.5} & \textbf{0.6} & \textbf{0.7} & \\
\midrule
1  & 52.25 & 46.94 & 40.73 & 31.26 & 20.17 & 38.27 \\
10 & 54.94 & 49.55 & \textbf{42.49} & 31.14 & 20.08 & 39.64 \\
16 & \textbf{56.42} & \textbf{50.54} & 42.48 & 32.35 & \textbf{21.17} & \textbf{40.59} \\
20 & 53.39 & 48.38 & 42.19 & \textbf{33.00} & 20.78 & 39.55 \\
30 & 50.27 & 45.54 & 38.30 & 29.64 & 18.83 & 36.52 \\
40 & 53.55 & 47.30 & 40.35 & 31.06 & 19.46 & 38.34 \\
\bottomrule
\end{tabular}
\end{table}

\subsubsection{FPN Levels}

In Tab~\ref{table:fpn_levels}, we show the effect of increasing or decreasing the number of feature pyramid levels in the network. The results show that six is the optimum number. Additional FPN layers beyond six will tend to increase the number of parameters for optimisation while not providing any additional benefit.

\subsubsection{Feature Matching Strategy}

To assess the efficacy of using Optimal Transport (OT) with the Sinkhorn Algorithm to align video features with adaptive prompts, we conducted ablation experiments in which OT was replaced with more straightforward distance metrics, precisely Euclidean distance and Hungarian distance. Our goal was to determine the impact of these substitutions on alignment performance and overall method effectiveness.

\subsubsection{Euclidean Distance} We replaced the OT metric with the Euclidean distance in the first variant. Here, the alignment between the refined video features $\{\vect{x'_1}, \vect{x'_2}, \dots, \vect{x'_T}\}$ and the adaptive prompts ${P}_k$ for each action category $k$ was performed directly using the Euclidean distance:

\[
d_{\text{Euc}}(\vect{U}, \vect{V_k}) = \sum_{t=1}^{T} \sum_{i=1}^{N} \| \vect{x'_t} - \vect{P_{ki}} \|^2
\]

In this formulation, the cost matrix $C_{ti}$ is defined as the squared Euclidean distance between video feature $\vect{x'_t}$ and prompt embedding $\vect{P_{ki}}$:

\[
C_{ti} = \| \vect{x'_t} - \vect{P_{ki}} \|^2
\]

The alignment process involves directly computing the sum of these distances without optimising a transport plan.

\subsubsection{Hungarian Distance} In the second variant, we utilised the Hungarian algorithm to find an optimal one-to-one matching between video features and prompts, minimising the overall distance. The cost matrix $C_{ti}$ is defined similarly to the Euclidean distance case, but the Hungarian algorithm ensures a unique assignment of each video feature to a prompt:

\begin{equation}
d_{\text{Hung}}(\vect{U}, \vect{V_k}) = \min_{\mat{T} \in \Pi} \sum_{t=1}^{T} \sum_{i=1}^{N} C_{ti} T_{ti}
\end{equation}

Here, $\Pi$ represents the set of all possible permutations that allow a one-to-one matching between the sets of video features and prompts. In Tab~\ref{table:alignment-method}, we show that OT outperforms both methods. 

The superior performance of OT can be attributed to several key factors:

\begin{itemize}
  
\item \textbf{Global Distribution Matching}: OT aligns the entire distribution of video features with the prompts distribution, considering the global structure and interdependencies within the data. In contrast, Euclidean distance considers each pair independently, which can lead to suboptimal alignments in the presence of complex feature distributions.

\item \textbf{Flexible Many-to-Many Matching}: OT allows for a many-to-many correspondence between video features and prompts, providing more flexibility in the alignment process. On the other hand, the Hungarian algorithm enforces a strict one-to-one matching, which may not capture the underlying relationships effectively, especially when the number of video features and prompts differ significantly.

\item \textbf{Entropic Regularization}: The Sinkhorn algorithm introduces entropic regularisation, promoting smoother and more stable solutions by avoiding challenging assignments. This regularisation helps mitigate the impact of noisy or outlier features, leading to more robust alignments.

\end{itemize}

\begin{table}[t]
\centering
\caption{Ablation on the number of FPN levels, evaluated on THUMOS'14. Performance peaks with a 5-level pyramid.}
\label{table:fpn_levels}
\begin{tabular}{@{}ccc@{}}
\toprule
\textbf{FPN Levels} & \textbf{mAP@0.5} & \textbf{Avg. mAP (\%)} \\
\midrule
1 & 25.82 & 26.16 \\
2 & 37.80 & 35.81 \\
3 & 39.10 & 36.58 \\
4 & 40.02 & 38.03 \\
5 & \textbf{43.06} & \textbf{40.46} \\
6 & 42.21 & 39.57 \\
7 & 41.56 & 38.92 \\
\bottomrule
\end{tabular}
\end{table}

\begin{table}[t]
\centering
\caption{Ablation on the prompt alignment strategy, evaluated on THUMOS'14. Our Optimal Transport (OT) approach significantly outperforms both hard-assignment (Kuhn-Munkres) and simple distance-based methods.}
\label{table:alignment-method}
\begin{tabular}{@{}lcc@{}}
\toprule
\textbf{Alignment Method} & \textbf{mAP@0.5} & \textbf{Avg. mAP (\%)} \\
\midrule
Euclidean Distance & 21.97 & 22.27 \\
Kuhn-Munkres (Hungarian) & 29.48 & 29.09 \\
\midrule
\textbf{Optimal Transport (OT)} & \textbf{43.06} & \textbf{40.46} \\
\bottomrule
\end{tabular}
\end{table}

\subsubsection{Visual Feature Embeddings}

To evaluate the effectiveness of adding motion information via optical flow, we also performed additional experiments using only the RGB embeddings, the optical flow embeddings, and RGB CLIP embeddings from a ViT-B-16 encoder, with results shown in Tab~\ref{tab:visual-embedding}. The results show that the CLIP embeddings perform better than the RGB from the I3D network $\uparrow 2.67$. This is because of the implicit alignment between the image and text encoder embeddings before temporal convolution. However, when combined with optical flow, the performance is improved by a large margin of $\uparrow 7.56$, demonstrating the enhanced classification ability of the network when we add additional temporal information.

%% file: sec/5_conclusion.tex
\section{Conclusion}
\label{conclusion}
 
 This work addressed a fundamental limitation in prevailing few-shot TAL methods: the inability of a single prompt vector to effectively model the compositional and dynamic nature of human actions from sparse data. We introduced PLOT-TAL, a framework that departs from this paradigm by modeling actions as a distribution of concepts, learned via an ensemble of diverse prompts. Crucially, we demonstrated that Optimal Transport serves not merely as a matching algorithm, but as a powerful structural regularizer that enforces prompt specialization, a key requirement for robust generalization in low-data regimes.

Through extensive experiments on THUMOS’14, EPIC-Kitchens, and ActivityNet 1.3, we established a new state-of-the-art in few-shot TAL without resorting to complex meta-learning schedules. Our analyses, particularly the significant performance gains at high IoU thresholds and the qualitative visualizations of prompt specialization, confirm that our method's success stems from learning a more precise and compositional representation of actions. By moving beyond mean-based representations towards structured, distributional alignments, our work opens a promising new direction for developing more generalizable and data-efficient models for video understanding.